\documentclass{article}

\usepackage[preprint]{neurips_2023}

\usepackage[utf8]{inputenc} 
\usepackage[T1]{fontenc}    
\usepackage{hyperref}       
\usepackage{url}            
\usepackage{booktabs}       
\usepackage{amsfonts}       
\usepackage{nicefrac}       
\usepackage{microtype}      
\usepackage{graphicx}
\usepackage{xcolor}         
\usepackage{algorithm}
\usepackage{fullpage}
\usepackage{amsmath,amsfonts,amsthm,amssymb}
\usepackage{algorithmic}
\usepackage{natbib}

\title{ContextWIN: Whittle Index Based Mixture-of-Experts Neural Model For Restless Bandits With Contextual Information Via Deep RL}

\author{%
  Zhanqiu Guo\\
  New York University\\
  \texttt{zg2238@nyu.edu} \\
  \And
  Wayne Wang\\
  New York University\\
  \texttt{yw5954@nyu.edu} \\
}

\begin{document}

\maketitle

\begin{abstract}
This study introduces ContextWIN, a novel architecture that extends the Neural Whittle Index Network (NeurWIN) model to address Restless Multi-Armed Bandit (RMAB) problems with a context-aware approach. By integrating a mixture of experts within a reinforcement learning framework, ContextWIN adeptly utilizes contextual information to inform decision-making in dynamic environments, particularly in recommendation systems. A key innovation is the model's ability to assign context-specific weights to a subset of NeurWIN networks, thus enhancing the efficiency and accuracy of the Whittle index computation for each arm. The paper presents a thorough exploration of ContextWIN, from its conceptual foundation to its implementation and potential applications. We delve into the complexities of RMABs and the significance of incorporating context, highlighting how ContextWIN effectively harnesses these elements. The convergence of both the NeurWIN and ContextWIN models is rigorously proven, ensuring theoretical robustness. This work lays the groundwork for future advancements in applying contextual information to complex decision-making scenarios, recognizing the need for comprehensive dataset exploration and environment development for full potential realization.
\end{abstract}

\section{Introduction}

The landscape of sequential decision-making in dynamic environments presents a complex yet fundamental challenge, particularly epitomized in the restless multi-armed bandit (RMAB) framework. Here, each decision or "arm" evolves overtime, with its state and reward potential changing based on past interactions. This dynamic nature is prominent in applications such as online advertising, where user responses continuously alter the effectiveness of each advertisement. A key hurdle in addressing RMAB problems is their computational complexity, driven by the ever-expanding state space of each arm.

Traditionally, the Whittle index policy has been a cornerstone in tackling RMAB problems, offering a heuristic that assigns an index to each arm based on its current state. However, the direct computation of Whittle indices for general RMABs remains an arduous task due to their inherent unobservability and computational complexity. Recent advances have led to the development of the Neural Whittle Index Network (NeurWIN), a machine-learning approach that approximates these indices for a broad range of RMAB problems. Yet, a singular NeurWIN model may not fully capture the contextual nuances present in complex, multi-dimensional RMAB scenarios.

In this paper, we propose an innovative extension to the NeurWIN model, incorporating a context-aware network that precedes a series of NeurWIN modules. This new architecture, which we refer to as ContextWIN, is designed to extract relevant contextual information from the environment and subsequently allocate weights to activate a subset of NeurWIN networks. Each NeurWIN module specializes in learning the Whittle index for individual arms, but it is the contextual layer that determines which of these experts to consult for a given decision. This mixture-of-experts approach allows ContextWIN to adaptively focus on the most pertinent arms in varying situations, enhancing both the efficiency and accuracy of decision-making in RMAB problems, especially the cold-starting recommendation here as our principle setting.

The structure of this paper is organized as follows: Section 2 reviews relevant literature, establishing the background and context for our work. Section 3 outlines our novel ContextWIN architecture and explains its integration with the NeurWIN modules. In Section 4, we detail the training algorithm and the implementation specifics of ContextWIN. Section 5 presents our experimental setup and findings, showcasing the practical effectiveness of ContextWIN in various RMAB scenarios. Finally, Section 6 concludes the paper with a summary of our contributions and a discussion on potential future research directions.

\section{Related Work}

The study in \cite{mate2020collapsing} delves into the complexities of the restless multi-armed bandit (RMAB) problem, elucidating the concept of the \textit{belief state}. This computed state is pivotal for estimating the Whittle index, a key tool in the optimization of RMABs. The paper \cite{mate2020collapsing} is instrumental in presenting a class of RMABs that are tractable via Whittle index policies.

The approach in \cite{mate2020collapsing} relies on the \textit{belief state}, derived from known transitional probabilities between the states of an arm. This contrasts with practical scenarios where such probabilities are not pre-determined but are learned through active interaction with the environment, posing a challenge for traditional models.

In contrast, \cite{nakhleh2022neurwin} proposes a novel framework, NeurWIN, that integrates reinforcement learning with deep learning (RL-DL) to accommodate the dynamic nature of arm behaviors. NeurWIN stands out by learning from posterior knowledge, adapting to environments where behavior is revealed progressively through interaction. Although NeurWIN offers a pragmatic solution for more realistic settings, it is worth noting that the framework does not extend theoretical performance guarantees. This recognition of NeurWIN's potential, juxtaposed with its theoretical limitations, underscores the ongoing evolution of solving RMAB problems in real-world applications.

\section{Problem Setting}

\subsection{Contextual Restless Bandit Problems in Recommendation System}

In a restless bandit problem, we are presented with \( N \) arms, each displaying restless behavior. For a contextual problem, each arm is accompanied by a fixed \( d \)-dimensional context vector. During each round \( t \), a control policy \( \pi \) assesses the state \( s_i[t] \) of each arm \( i \) and activates \( K \) of them. The arms that are chosen for activation are termed 'active', while the rest are deemed 'passive'. The action of the policy on arm \( i \) at round \( t \) is represented by \( a_i[t] \), where \( a_i[t] = 1 \) indicates an active arm, and \( a_i[t] = 0 \) signifies a passive one. The reward generated by arm \( i \) at round \( t \), denoted as \( r_i[t] \), follows a stochastic distribution: \( R_{i,\text{act}}(s_i[t]) \) if the arm is active, and \( R_{i,\text{pass}}(s_i[t]) \) if it is passive. Depending on the arm's activity status, its state transitions according to either \( P_{i,\text{act}}(s_i[t]) \) or \( P_{i,\text{pass}}(s_i[t]) \) in the next round. The overarching aim of the control policy is to maximize the expected cumulative discounted reward, formulated as \( \mathbb{E}_\pi\left[\sum_{t=0}^{\infty} \sum_{i=1}^{N} \beta^t r_i[t]\right] \), where \( \beta \) is the discount factor.

A user case cold start recommendation problem refers to when a new user enrolls in the system and for a certain period of time the recommender has to provide recommendations without relying on the user's past interactions since none has occurred yet.  It fits in the contextual restless bandit problem as a decision problem for the recommendation system to select \(K\) items from a total of  \(N\) items in each round \(t\) with a context of information for all these items expecting to maximize the expected cumulative discounted reward for \(r\) being the clicking rate here. As users' behavior and interests could vary over time, it is a restless bandit problem as the possibility of clicking will keep varying over time for all items no matter they are clicked or not. 

A control policy essentially maps the state vector \( (s_1[t], s_2[t], \ldots, s_N[t]) \) to the action vector \( (a_1[t], a_2[t], \ldots, a_N[t]) \). The complexity arises from the fact that the space of possible inputs is exponential in \( N \). With \( |S| \) possible states for each arm, there are \( |S|^N \) potential input combinations, a dilemma often referred to as the curse of dimensionality, rendering the search for an optimal control policy a formidable task. 

\subsection{Whittle Index}

In the study of index policies for sequential decision-making problems, particularly in the context of multi-armed bandit models, the Whittle index offers a compelling solution to the challenge of dimensionality. It functions by decomposing the decision space and computing an index, denoted by \( W_i(s_i[t]) \), for each arm \( i \) based on its present state. This index is then employed to select the \( M \) arms with the highest values to be activated. It's important to recognize that the state of one arm is not contingent on the states of others, making the task of learning the Whittle index for restless arms a subsidiary to determining the optimal control policy for such bandits.

The efficacy of an index policy is largely contingent on the design of its index function, \( W_i(\cdot) \), and the Whittle index stands out due to its robust theoretical underpinnings. For simplification, when addressing only a single arm, the subscript \( i \) is omitted.

Delving into a scenario with a solitary arm, we consider an activation policy that adjudicates whether to activate it each round \( t \). This policy necessitates a payment of \( \lambda \) for activation at every instance. The objective is to maximize the expected net reward over time, encapsulated by the equation \( E\left[\sum_{t=0}^{\infty} \beta^t(r[t] - \lambda[t])\right] \), where \( \beta \) is the discount factor. The optimal policy can thus be characterized by the states in which it would activate the arm for a given \( \lambda \), denoted as \( S(\lambda) \). Logically, a higher activation cost would reduce the probability of the arm being activated, leading to a monotonic decrease in the set \( S(\lambda) \). An arm meeting these conditions is termed indexable.

\textbf{Definition 1 (Indexability).} An arm is said to be \textit{indexable} if \( S(\lambda) \) decreases monotonically from the set of all states to the empty set as \( \lambda \) increases from \( -\infty \) to \( \infty \). A restless bandit problem is said to be indexable if all arms are indexable.

\textbf{Definition 2 (Strong Indexability).} A system is said to be strongly indexable if every arm in the system is indexable and a Whittle index can be assigned to each state of each arm. Additionally, for a given arm, if its state has a higher index than another state, it must always be at least as good to activate the arm in that state than in the other, regardless of the activation cost. 

\textbf{Definition 3 (The Whittle Index).} If an arm is indexable, then its Whittle index of each state \( s \) is defined as \( W(s) := \sup\{\lambda : s \in S(\lambda)\} \).

It is crucial to prove the strong indexability for our system to ensure the exsitence of an optimal neural network for predicting whittle index as described in  \cite{nakhleh2022neurwin}.  \cite{mate2020collapsing} proposes to use a \textit{belief state}, the probability of being in each latent state, to analyze partially observable RMABs. In settings with limited observability, belief-state Markov Decision Processes (MDPs) exhibit organized chain-like structures. These structures are beneficial for the analytical process. Specifically, the pivotal information determining our belief about an arm being in state 1 is the elapsed time since the arm was last activated, coupled with the observed state \( \omega \) at that moment. Consequently, these belief states can be systematically organized into two sequences, or "chains", with length \( T \), corresponding to each observation \( \omega \). Figure 1 illustrates the belief state chains under a consistently passive policy. We denote \( b_\omega(u) \) as the belief state, signifying the probability that the state is 1, given that the observation \( \omega \in \{0, 1\} \) was made when the agent last interacted with the process \( u \) days prior. It is important to note that \( b_\omega(u) \) also represents the expected reward linked to that belief state. Let \( \mathcal{B} \) denote the complete set of belief states.
\begin{figure}[h]
\centering
\includegraphics[width=0.8\textwidth]{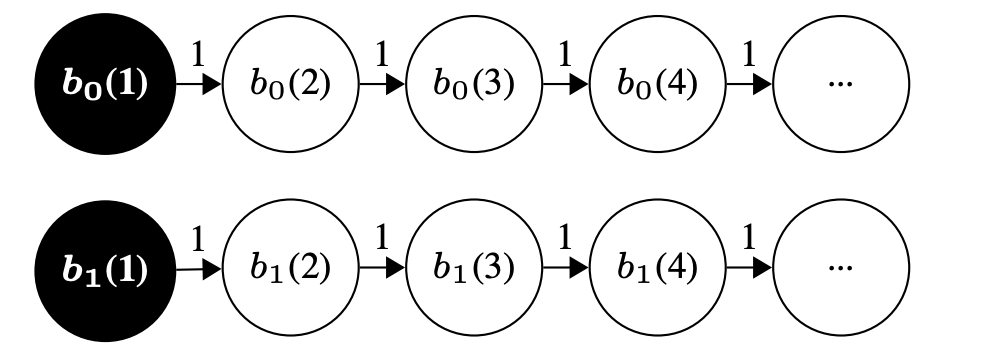}
\caption{Belief-state MDP under the policy of remaining passive. Each chain represents a belief state for an observation \( \omega \) within \( \{0, 1\} \), with the initial state highlighted. Transitions between belief states occur deterministically along the chains.}
\end{figure}

Thus the policy can be modeled as a threshold policy with a belief state threshold $b_{th}$. 

\textbf{Definition 4 (Threshold Policies).} A policy is a reverse threshold policy if there exists a threshold $b_{th}$ such that $\pi(b) = 1$ if $b > b_{th}$ and $\pi(b) = 0$ otherwise.  

With definitions in hand, we can prove the strong indexability of our system in \textbf{Theorem 1}. 

\textbf{Lemma 1}
Given a monotonically increasing belief state, the threshold at which the policy changes from passive to active is non-decreasing with respect to the penalty \(\lambda\).

\textbf{Theorem 1}
The system is strongly indexable under the assumption of a monotonically increasing belief state with respect to the reverse threshold policy.

This result is essential for proving the existence of an optimal neural network for learning the accurate whittle index in our recommendation system application. The detailed proof is in \textbf{Appendix A}. 

\subsection{NeurWIN}

\cite{nakhleh2022neurwin} presents a deep-RL algorithm that trains neural networks to predict the Whittle indices, NeurWin. Since the Whittle index of an arm is independent of other arms, NeurWIN trains one neural network for each arm independently; therefore, we will use the calculation for one arm to demonstrate the structure of NeurWin.
NeurWin has a simple structure. Let us assume that we have access to a simulator for an individual arm as well as a parameterized neural network denoted as \( \theta \). This enables the construction of a virtual environment for the arm, which also incorporates an activation cost denoted by \( \lambda \), as depicted in Figure 1. At every iteration \( t \), this environment processes the real-valued function \( f_\theta(s[t]) \) as its input.

The procedure begins by passing the input through a step function to determine the action \( a[t] = 1 \) if \( f_\theta(s[t]) > \lambda \), where \( 1(\cdot) \) symbolizes the indicator function. Subsequently, \( a[t] \) is channeled into the arm's simulator, yielding the reward \( r[t] \) and the subsequent state \( s[t + 1] \). The environment then generates the net reward \( r[t] - \lambda a[t] \) along with the new state \( s[t + 1] \). This setup is referred to as the environment \( \text{Env}_\lambda \). In this context, the neural network is essentially acting as a controller for \( \text{Env}_\lambda \). The ensuing corollary emerges directly from the aforementioned environment.
\begin{figure}[h]
\centering
\includegraphics[width=0.8\textwidth]{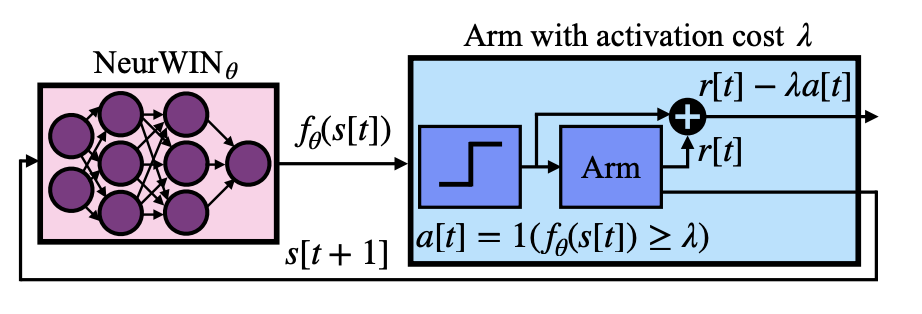}
\caption{An illustrative motivation of NeurWIN. The neural network, represented here, interfaces with the arm simulator, processing the inputs and outputs according to a specified activation cost \( \lambda \).}
\end{figure}

\subsection{Mixture of Experts}

The Mixture of Experts (MoE) model, introduced by Jacobs et al. (1991), represents a sophisticated approach to modular neural network design. It encompasses a collection of expert models \( E_1, E_2, \ldots, E_n \) and a gating network \( G \) that dynamically allocates the task to the appropriate expert based on the input features \( x \). Each expert \( E_i \) is typically a neural network trained to specialize in a specific region of the input space, while the gating network is a probabilistic model that outputs weights \( w_1, w_2, \ldots, w_n \) corresponding to the relevance of each expert for a given input.

The overall output \( y \) of an MoE model for an input \( x \) is computed as a weighted sum of the experts' outputs:

\[
y(x) = \sum_{i=1}^{n} w_i(x) \cdot E_i(x)
\]

where \( w_i(x) \) are the weights provided by the gating network, signifying the probability that expert \( E_i \) is the most suitable for the current input. The weights are normalized such that \( \sum_{i=1}^{n} w_i(x) = 1 \), ensuring a proper probabilistic mixture.
\begin{figure}[h!]
\centering
\includegraphics[width=0.8\textwidth]{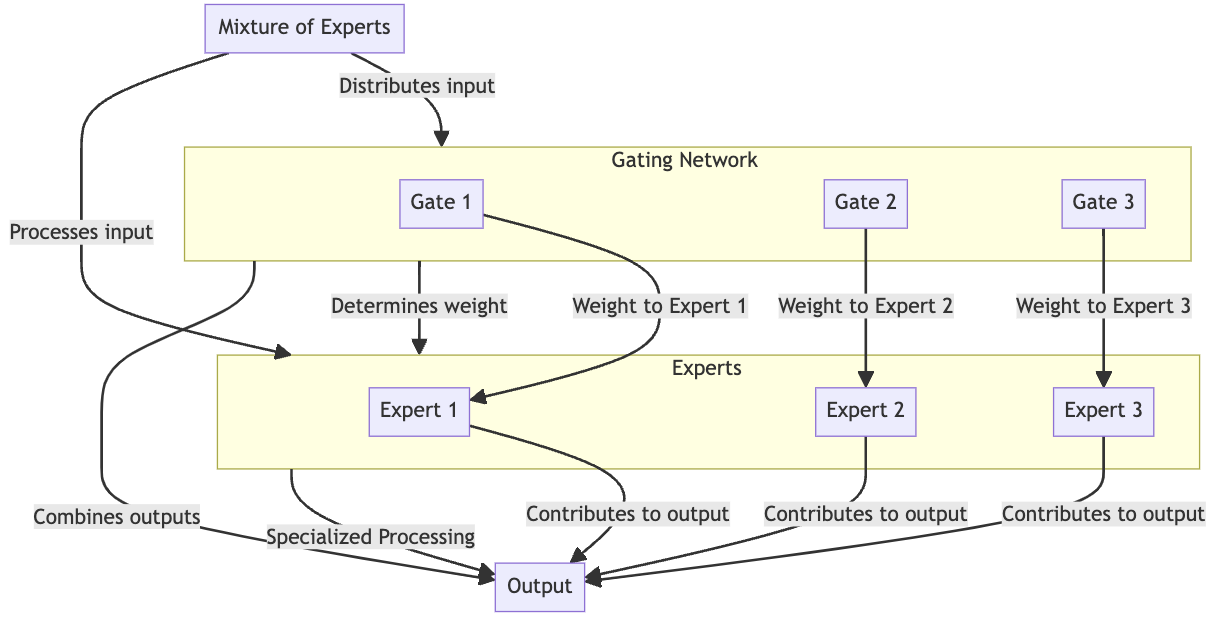}
\caption{Mixture of Experts}
\end{figure}

The training of MoE models is executed via an optimization process that involves adjusting both the expert networks and the gating network to minimize a global objective function, often a loss function \( L \) indicative of the prediction error. The typical choice of optimization algorithm is gradient descent or its variants, with backpropagation employed to compute the necessary gradients.

\subsection{Problem Statement}

The central aim of this paper is to derive a data-efficient low-complexity index algorithm for restless bandit problems by training a competitive mixture-of-experts structure (MoE) that predicts the whittle index of each arm instead of training individual networks for each item. In contextual bandit setting, we proposed a MoE structure ContextWIN model that takes context $x$ as the input for the gating network and the state $s$ as the input for each expert network, outputting a weighted result \(\sum_{i=1}^{n} G_i(x) f_{\theta^{(i)}}(s_0)\) as an estimation of the whittle index, where $w$ is all the weights of the mixture-of-experts structure. 

NeurWIN\cite{nakhleh2022neurwin} has proved that an $\epsilon$-optimal neural network can ensure $\gamma$-accurate whittle index output with each arm indexable, our goal is to prove that a single expert neural network converges. With the convergence of each expert and the gating network (gating network also pretrainable), we can thus ensure the convergence of the whole MoE model.

Prove that single experts NeurWIN converge to \(\gamma\) accurate network and the MoE model also converges based on a certain number of experts in a pre-trained setting. 

\section{ContextWIN}

In this section, we present ContextWin. ContextWIN is a deep-RL and trains a mixture-of-experts structure to predict the Whittle indices. Since the behavior, and thus the Whittle index, of an arm is independent of other arms, we will present how ContextWIN learns to calculate the Whittle index of one arm. 

\subsection{Conditions for Whittle-Accurate}

The procedure of the ContextWIN algorithm is as follows.
\begin{itemize}
    \item At each time step $t$, each expert takes in the current state, $s[t]$ as input and outputs an estimation, $f_{{\theta}^i}(s[t])$ where the $i$ denotes the $i$the expert.
    \item The gating network then takes in all the $f_{{\theta}^i}(s[t])$ and the context $x$ as input, assigning a weight, $G_i(x)$ for each of the top-k results calculated by experts. The gating network outputs the weight estimation of Whittle index, \(\sum_{i=1}^{k} G_i(x) f_{\theta^{(i)}}(s)\)
    \item The weight estimation of Whittle index is fed into a step function $1( \sum_{i=1}^{k} G_i(x) f_{\theta^{(i)}}(s)\geq \lambda)$, where \( 1(\cdot) \) is the indicator function.
    \item $a(t)$ represents whether an action is taken for the arm. Therefore, we can understand $a(t)$ as what we fed into the arm (the environment).A reward $r(t)$ and a net reward $r(t) - \lambda a(t)$ is outputted by the arm. 
    \item the arm also produces the state at the next time stamp, $s[t+1]$.
\end{itemize}
    The entire procedure can be visualized in Figure 4.

\begin{figure}[H]
\centering
\includegraphics[width=0.8\textwidth]{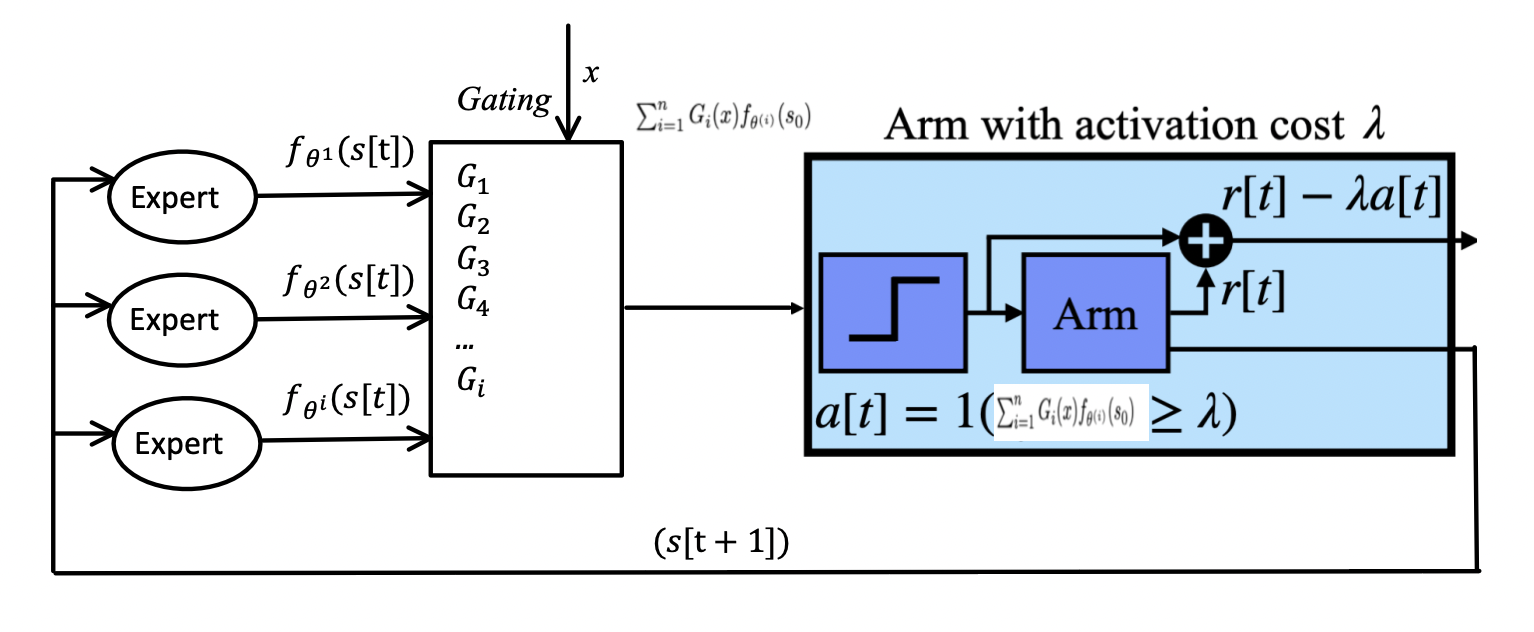}
\caption{ContextWIN}
\end{figure}
    
In \cite{nakhleh2022neurwin}, it is defined that a neural network with parameters $\theta$ is said to be $
\gamma -$Whittle-accurate if $|f_{\theta}(s)-W(s)|\le \gamma$, where $W(s)$ is the actual Whittle index of the arm and $f_{\theta}(s)$ is the estimated Whittle index.\\
In addition to the definition above, \cite{nakhleh2022neurwin} proves that if the arm is strongly indexable, then for any \(\gamma > 0\), there exists a positive \(\epsilon\) such that any \(\epsilon\)-optimal neural network controlling \(\text{Env}(\lambda)\) is also \(\gamma\)-Whittle-accurate. Since we have established that our arm is strongly indexable in Appendix A and our mixture-of-expert can be interpreted as a complex neural network, the conclusion for our structure. In Appendix B, we advance to prove that our model converges, which is a major part of this paper.

\subsection{Training of ContextWIN}
We obtain the structure of training from \cite{nakhleh2022neurwin}, therefore, we define our objective function based on \cite{nakhleh2022neurwin} and our mixture-of-experts model:  \( \sum_{i=1}^{n} G_{\omega}(x)^{(i)}\sum_{s_0, s_1} \tilde{Q}_{\theta^{(i)}}(s_1, \lambda = f_{\theta^{(i)}}(s_0))\), where the estimated index $f_{\theta^{(i)}}(s_0))$ set as the environment's activation cost for the $i$th expert.\\
We define \(G(x) = \text{TopK}(\text{softmax}(W_g (x) + \epsilon)) : \mathbb{R}^D \rightarrow \mathbb{R}^N
\), \(\epsilon \sim \mathcal{N}(0, \frac{1}{M^2})\), where $\epsilon$ serves to loose our choice of experts.\\
For the gradient to exist, we require the output of the arm(environment) to be differentiable concerning the input \(f_\theta(s[t])\) (at this step, we do not address the effect of gating for easier computations). To fulfill the requirement, we replace the step function in Fig. 1 with a sigmoid function,
\[
\sigma_m(f_\theta(s[t]) - \lambda) := \frac{1}{1 + e^{-m(f_\theta(s[t])-\lambda)}}
\]
Where \( m \) is a sensitivity parameter. The environment then chooses \( a[t] = 1 \) with probability \( \sigma_m(f_\theta(s[t]) - \lambda) \), and \( a[t] = 0 \) with probability \( 1 - \sigma_m(f_\theta(s[t]) - \lambda) \). We call this our environment \( \text{Env}^*(\lambda) \).
\begin{algorithm}[H]
\caption{ContextWIN Training}
\begin{algorithmic}
\REQUIRE Expert parameters $\theta$, expert models $\{\theta_1, \theta_2, \ldots, \theta_M\}$, gating network parameters $\omega$, discount factor $\beta \in (0, 1)$, learning rate $L_b$, sigmoid parameter $m$, mini-batch size $R$.
\ENSURE Trained neural network parameters $\theta^*$, trained expert models $\{\theta_1^*, \theta_2^*, \ldots, \theta_M^*\}$, trained gating network parameters $\omega^*$.

\FOR{each mini-batch $b$}
    \STATE Choose two states $s_0$ and $s_1$ uniformly at random, and set $\lambda \leftarrow f_\theta(s_0)$ and $\overline{G}_b \leftarrow 0$
    \FOR{each episode $e$ in the mini-batch}
        \STATE Set the arm to initial state $s_1$, and set $h_e \leftarrow 0$
        \STATE Calculate gating weights: $g = \text{GatingNetwork}(\omega)$
        \FOR{each round $t$ in the episode}
            \STATE Choose $a[t] = 1$ w.p. $\sigma_m(f_\theta(s[t]) - \lambda)$, and $a[t] = 0$ w.p. $1 - \sigma_m(f_\theta(s[t]) - \lambda)$
            \IF{$a[t] = 1$}
                \FOR{each expert model $i = 1$ to $M$ with \(g_i\) in  Top K}
                    \STATE $h_e^i \leftarrow h_e^i + g^i \nabla_{\theta_i} \sigma_m(f_{\theta_i}(s[t]) - \lambda)$
                \ENDFOR
            \ELSE
                \FOR{each expert model $i = 1$ to $M$ with \(g_i\) in  Top K}
                    \STATE $h_e^i \leftarrow h_e^i + g^i \nabla_{\theta_i} (1 - \sigma_m(f_{\theta_i}(s[t]) - \lambda))$
                \ENDFOR
            \ENDIF
        \ENDFOR
        \STATE $G_e \leftarrow \text{empirical discounted net reward in episode } e$
        \STATE $\overline{G}_b \leftarrow \overline{G}_b + G_e / R$
    \ENDFOR

    \STATE $L \leftarrow \text{expert learning rate in mini-batch}$
   \STATE $L_g \leftarrow \text{gating learning rate in mini-batch}$
    \FOR{each expert model $i = 1$ to $M$}
        \STATE Trough gradient ascent $\Delta\theta_i  \leftarrow\frac{L}{R}  \sum_e (G_e - \overline{G}_b) h_e^i$ 
        \STATE Update expert model parameters through gradient ascent $\theta_i \leftarrow \theta_i + \Delta\theta_i$
    \ENDFOR
     \STATE Update gating network parameters through gradient ascent $\omega \leftarrow \omega + L_g \sum_i \theta_i \nabla_{\omega} g_i$
\ENDFOR
\end{algorithmic}
\end{algorithm}

Our complete pseudocode for training is in Algorithm 1. 
The procedure is designed to train a collection of expert neural networks in tandem with a gating network. The training involves numerous mini-batches, each consisting of \( R \) episodes, wherein two states \( s_0 \) and \( s_1 \) are randomly selected to initialize the system. Leveraging a gating network parameterized by \( \omega \), the algorithm dynamically assigns weights to the expert models based on the current state, optimizing the ensemble’s response to the environment \( \text{Env}^* \) as a function of the initial condition \( f_{\theta}(s_0) \).

Within each episode \( e \) of a given mini-batch, the gating network is applied to determine the contributions of the top \( K \) expert models. These experts, parameterized by \( \{\theta_1, \theta_2, \ldots, \theta_M\} \), then engage in a collaborative effort to inform the action selection process. The sequence of actions \( (a[t_1], a[t_2], \ldots) \) and corresponding states \( (s[t_1], s[t_2], \ldots) \) observed furnish the data required for the gradient calculations through back propagation, signified by \( h^i_e \) for each expert model \( i \).

For each action decision, whether to act (\( a[t] = 1 \)) or not (\( a[t] = 0 \)), gradients are accumulated across the top \( K \) experts based on their gated influence. Following the conclusion of all episodes in the mini-batch, the empirical discounted net reward for each episode, \( G_e \), is computed. These rewards are then averaged to form a bootstrap baseline \( G_b \), which serves as a reference for updating the gradient ascent.

The update procedure entails adjusting the gating network parameters \( \omega \) as well as the parameters for each expert model \( \theta_i \) through gradient ascent, ensuring that the influence of each episode is proportionate to its net reward offset \( G_e - G_b \). Through meticulous selection of the learning rate and step size, the integrated system of the gating network and expert models is fine-tuned, to follow action sequences that culminate in superior empirical discounted net rewards.

\subsection{Convergence of ContextWIN}

To prove the convergence of the whole MoE model, we first investigate the convergence quality of a single basic NeurWIN expert model as described in \cite{nakhleh2022neurwin}.  The detailed proof can be found in Appendix B. 

In \cite{nakhleh2022neurwin}, the author has claimed the smoothness of NeurWIN's loss function, we are given a

\textbf{Definition 5 (\(\beta\)-smooth).} A function $J$ corresponding with parameters $\theta$ is called $\beta$-smooth if: 
\[\| \nabla J(\theta) - \nabla J(\theta') \|_2 \leq \beta \| \theta - \theta' \|_2 \]
\[\|J(\theta) - J(\theta') - \nabla J(\theta')^T (\theta - \theta')\|_2 \leq \frac{\beta}{2} \| \theta - \theta' \|_2^2, \forall \theta, \theta'\]

NeurWIN's loss function is $\beta$-smooth as an estimation of \( J(\theta) \) representing the expected return when following a whittle index policy denoted as \( \pi_\theta \) with a smooth $\beta$-smooth sigmoid activation function. 

As a policy gradient method based on REINFORCE, NeurWIN fits into the policy gradient theorem described as \textbf{Theorem 2}.

\textbf{Theorem 2 (Policy Gradient Theorem)}. 
\[
\nabla_\theta J(\theta) = \mathbb{E}_{\tau \sim \pi_\theta} \left[ \sum_{t=0}^{T} \nabla_\theta \log \pi_\theta(a_t | s_t) \cdot R_t \right]
\]
The policy \( \pi_\theta(a_t | s_t) \), is a probability distribution over actions \( a_t \) given the states \( s_t \). The gradient of the logarithm of this policy with respect to its parameters \( \theta \), is expressed as \( \nabla_\theta \log \pi_\theta(a_t | s_t) \). Furthermore, \( R_t \) signifies the return, which is the cumulative reward from time \( t \) onwards. The expectation over the distribution of trajectories \( \tau \), represented as \( \mathbb{E}_{\tau \sim \pi_\theta} \), encompasses sequences of states and actions under the policy \( \pi_\theta \). Finally, \( T \) defines the time horizon of the trajectory under consideration.

The theorem indicates that the direction of the greatest increase in expected return is given by the expectation of the product of the gradient of the policy's log-probability concerning its parameters and the return. We thus need to 

\textbf{Lemma 2 (Unbiased Gradient Estimate).} 
The expectation of the step update of NeurWIN parameter \( \mathbb{E}\left[\frac{1}{R}\sum_e (G_e - \bar{G}_b) h_e \right]\) is an unbiased estimator of the gradient $\nabla_\theta J(\theta)$. 

\(\mathbb{E}\left[\frac{1}{R}\sum_e h_e \right]\) is an unbiased estimator of the gradient $\nabla_\theta J(\theta)$ according to the definitions. Multiplying the coefficient \((G_e - \bar{G}_b)\) is a technique to reduce variance that won't affect the unbiasedness. 

\textbf{Lemma 3 (Reducing Gradient Estimate Variance).}

The variance of the gradient estimate $Var\left[\frac{1}{R}\sum_e (G_e - \bar{G}_b) h_e \right]$ is reduced by the mini-batch size $R$. 

With the basic property of variance, we have: 
\begin{align*}
    \text{Var}\left( \frac{1}{R} \sum_{e=1}^{R} (G_e - \bar{G}_b)h_e \right) &= \frac{1}{R^2} \sum_{e=1}^{R} \text{Var}((G_e - \bar{G}_b)h_e)\\
    &= \frac{1}{R^2} \cdot R \cdot \sigma^2 \\
    &= \frac{\sigma^2}{R}
\end{align*}
Furthermore, according to the Central Limit Theorem, as the mini-batch size $R$ becomes large, the distribution of the sum
\[
\frac{1}{R} \sum_{e=1}^{R} (G_e - \bar{G}_b)h_e
\]
approaches a normal distribution with a mean equal to the expected value of the summands and variance equal to $\frac{\sigma^2}{R}$. This implies that for large $R$, the mini-batch gradient estimate becomes more stable and reliable for gradient ascent updates.

This shows that the variance of the gradient estimate is indeed reduced by the mini-batch size $R$, and as $R$ increases, the variance decreases. This is consistent with the intuition that averaging over more samples should result in a more stable estimate.

\textbf{Theorem 3 (Convergence of NeurWIN).}

If \( J(\theta) \) is \( \beta \)-smooth, and we run mini-batch Stochastic Gradient Ascent with the update rule:
\[
\theta_{b+1} = \theta_b + \frac{L}{R} \sum_{e=1}^{R} (G_e - \bar{G}_b) h_e,
\]
where \( R \) is the mini-batch size, \( L \) is the learning rate for mini-batch \( b \), \( G_e \) is the empirical discounted net reward for an episode \( e \), \( \bar{G}_b \) is the average of rewards \( G_e \) over the mini-batch, and \( h_e \) is the eligibility trace for an episode \( e \), such that:
\[
\mathbb{E}\left[\sum_{e=1}^{R} (G_e - \bar{G}_b) h_e\right] = \nabla_\theta J(\theta_b), \quad \text{and} \quad \mathbb{E}\left[\left\| \frac{1}{R}\sum_{e=1}^{R} (G_e - \bar{G}_b) h_e \right\|_2^2\right] \leq \frac{\sigma^2}{R},
\]
then, taking \( L \) proportional to \( \sqrt{\frac{RM}{\beta \sigma^2 B}} \):
\[
\mathbb{E}\left[ \frac{1}{B} \sum_{b=1}^{B} \|\nabla_\theta J(\theta_b)\|_2^2 \right] \leq \mathcal{O}\left( \frac{\sigma\sqrt{R\beta} }{\sqrt{B}} \right),
\]
where \( B \) is the total number of mini-batches.

This could be derived by using the property of $\beta$-smooth and unbiased estimator. The detailed proof is in Appendix B. Thus we have shown the convergence of a single expert model. With a similar setting, we can prove that the gating network also converges given expert models fixed. Thus the training convergence of ContextWIN has been proved. With the gating network pre-trainable and supporting transfer-learning with the assumption that it mainly captures context information that is not specialized to each individual user, we could freeze the gating network or give it a penalty for changing, making the gating network changes insignificant. Thus the convergence of training ContextWIN model in practice could have an upper bound of \(\mathbb{E}\left[ \frac{1}{B} \sum_{b=1}^{B} \|\nabla J\|_2^2 \right] \leq \mathcal{O}\left( \frac{N\sigma\sqrt{R\beta} }{K\sqrt{B}} \right)\) with $J$ representing the expected return for ContextWIN, $N$ representing the total number of items, and $K$ representing the expert number. 

\section{Conclusion}

\subsection{Summary of Work}

In summary, the ContextWIN model embodies a fusion of expert insights within a reinforcement learning framework. Three major contributions define our work: firstly, we have established indexability within the specific use case of recommendation systems. Secondly, we have rigorously demonstrated the convergence of the NeurWIN model. Thirdly, and crucially, we have extended this proof of convergence to the ContextWIN model, enhancing its theoretical foundation.

It is important to note that these findings are preliminary and that ContextWIN's performance has not yet been evaluated against a comprehensive test dataset or within a fully developed testing environment. This recognized limitation underscores the need for further research to fully ascertain the capabilities and potential of ContextWIN.

Future efforts will be directed towards gathering a more robust dataset and creating a detailed evaluation environment to better understand the practical applications of the model. This endeavor will help to refine ContextWIN's utility and ensure its readiness for broader implementation. Moving forward, the focus will be on careful validation and thoughtful integration of feedback to enhance the model's performance in real-world scenarios.

\subsection{Report}

Our initial objective was to enhance the online algorithm proposed by Mate et al. (2020) by incorporating contextual elements. We hypothesized that adding context would not only facilitate convergence but also improve the overall performance of the algorithm. In exploring solutions, we considered the application of contextual bandits with Whittle index, aiming to leverage context for grouping similar arms, thereby transforming the problem into a standard Restless Multi-Armed Bandit (RMAB) challenge. However, this method did not fully capitalize on the context in learning the state transition behaviors of individual arms. 

To address this, we shifted our focus to integrating context information directly into the policy optimization process. This led us to the policy gradient method, specifically REINFORCE, and subsequently to the development of NeurWIN, which employs a neural network to compute the Whittle index. We identified the recommendation system as a fitting scenario for this contextual bandit problem. The challenge, however, was the sheer number of items (or arms) in the system, making it impractical to assign a separate NeurWIN model to each item. 

Recognizing that individual behaviors exhibit patterns that are not entirely independent or unique, we explored the potential of transfer learning and pre-training to expedite the convergence of each individual's learning process. Furthermore, it became apparent that an algorithm efficient in data utilization was essential—one that could update an individual's preferences across all items based on a single action. This realization led us to integrate a Mixture of Experts (MoE) structure, where a gating network comprehends the context and assigns weights to the calculations of each expert. With such a structure, it is possible to set the expert number far lower than the item (arm) number and have a faster rate of convergence by network information sharing. 

In our work, we also addressed a gap identified by Nakhleh et al. (2022), where only the existence of an optimal network was established. We dedicated substantial effort to proving the convergence of both the original NeurWIN structure and our ContextWIN model.

Efforts to create a simulating environment to test the performance of ContextWIN were challenged by limitations in the dataset and time constraints. A significant hurdle was the lack of prior knowledge about the environment's response, a crucial element for effective reinforcement learning training. Despite these challenges, our work lays the groundwork for future advancements in the application of contextual information in complex decision-making scenarios.

\bibliographystyle{apalike}
\bibliography{reference}

\newpage
\section*{Appendix}

\section*{A. Proof of Indexability}

\subsection*{Lemma 1}
Given a monotonically increasing belief state, the threshold at which the policy changes from passive to active is non-decreasing with respect to the penalty \(\lambda\).

Suppose \(b^{th}(\lambda)\) represents the threshold for a given penalty \(\lambda\). If the belief is above this threshold, the arm becomes active. Assume, for a contradiction, that there exist penalties \(\lambda_1 < \lambda_2\) such that \(b^{th}(\lambda_1) > b^{th}(\lambda_2)\). This would imply that an arm that is active for the lower penalty \(\lambda_1\) could potentially become passive for a higher penalty \(\lambda_2\), contradicting the assumption that the belief state monotonically increases with the penalty. Therefore, our assumption must be false, and \(b^{th}(\lambda)\) must be non-decreasing.

\subsection*{Theorem 1}
The system is strongly indexable under the assumption of a monotonically increasing belief state with respect to the reverse threshold policy.

Let us consider two penalties \(\lambda_1\) and \(\lambda_2\) such that \(\lambda_1 < \lambda_2\). According to Lemma 1, the threshold for activation under penalty \(\lambda_1\) is less than or equal to that under penalty \(\lambda_2\), i.e., \(b^{th}(\lambda_1) \leq b^{th}(\lambda_2)\).

Now, let us consider a belief state \(b\). If \(b \geq b^{th}(\lambda_1)\), then the arm is active for penalty \(\lambda_1\). Since \(b^{th}(\lambda_1) \leq b^{th}(\lambda_2)\), it follows that \(b \geq b^{th}(\lambda_2)\), and hence the arm must also be active for penalty \(\lambda_2\). This satisfies the definition of strongly indexable, which states that if an arm is active for a lower penalty, it remains active for any higher penalty.

This completes the proof that the system is strongly indexable under the given assumptions.

\section*{B. Proof of Convergence of ContextWIN}

\subsection*{Theorem 3: Convergence of NeurWIN}

Given the \( \beta \)-smoothness of \( J(\theta) \), we have the following inequality for all \( \theta \):
\[
\|J(\theta') - J(\theta) - \nabla J(\theta)^T(\theta' - \theta)\|_2  \leq \frac{\beta}{2} \|\theta' - \theta\|^2, \quad \forall \theta, \theta'
\]
Thus we can obtain an upper bound for \(\nabla J(\theta)^T(\theta' - \theta)\) as following: 
\[
\nabla J(\theta)^T(\theta' - \theta) \leq J(\theta') - J(\theta) + \frac{\beta}{2} \|\theta' - \theta\|^2
\]
Substituting the update rule into the smoothness inequality, we get:
\begin{align*}
\nabla J(\theta_b)^T(\theta_{b+1} - \theta_b) &\leq J(\theta_{b+1}) - J(\theta_b)  + \frac{\beta}{2} \|\theta_{b+1} - \theta_b\|^2 \\
\frac{L}{R} \nabla J(\theta_b)^T \sum_{e=1}^{R} (G_e - \bar{G}_b) h_e &\leq J(\theta_{b+1}) - J(\theta_b)+ \frac{\beta L^2}{2} \left\| \frac{1}{R}\sum_{e=1}^{R} (G_e - \bar{G}_b) h_e \right\|_2^2
\end{align*}
Taking the expectation and using the unbiasedness of the gradient estimate, we have:
\[
 \frac{L}{R} \|\nabla J(\theta_b)\|_2^2\leq  \mathbb{E}[J(\theta_{b+1}) - J(\theta_b)]+ \frac{\beta L^2 \sigma^2}{2R}.
\]
Summing over all mini-batches from \( b = 1 \) to \( B \) and dividing by \( B \), we obtain:
\begin{align*}
 \frac{1}{B}\sum_{b=1}^B \|\nabla J(\theta_b)\|_2^2 &\leq \frac{R}{BL}\mathbb{E}[J(\theta_B) - J(\theta_1)] + \frac{\beta L \sigma^2}{2}\\
 &\leq \frac{RM}{BL} + \frac{\beta L \sigma^2}{2}
\end{align*}
Where $M$ is a constant that bounds $\mathbb{E}[J(\theta_B) - J(\theta_1)]$ with $J$'s $\beta$-smooth property. Choosing \( L \) appropriately, for example, setting \( L \) proportional to \( \sqrt{\frac{RM}{\beta \sigma^2 B}} \), we can ensure that:
\[
\frac{1}{B} \sum_{b=1}^{B} \|\nabla J(\theta_b)\|_2^2 \leq \mathcal{O}\left( \frac{\sigma\sqrt{R\beta} }{\sqrt{B}} \right).
\]
This concludes the proof, showing that the mini-batch Stochastic Gradient Ascent converges under these conditions and thus demonstrates the convergence of individual expert NeurWIN model.

\end{document}